\documentclass[letterpaper]{article} 
\usepackage[draft]{aaai2026}  
\usepackage{times}  
\usepackage{helvet}  
\usepackage{courier}  
\usepackage[hyphens]{url}  
\usepackage{graphicx} 
\usepackage{natbib}  
\usepackage{caption} 
\usepackage{algorithm}
\usepackage{algorithmic}
\usepackage{newfloat}
\usepackage{listings}
\DeclareCaptionStyle{ruled}{labelfont=normalfont,labelsep=colon,strut=off} 
\floatstyle{ruled}
\newfloat{listing}{tb}{lst}{}
\floatname{listing}{Listing}

\usepackage{booktabs}
\usepackage{amsmath}

\usepackage{xspace}
\makeatletter
\DeclareRobustCommand\onedot{\futurelet\@let@token\@onedot}
\def\@onedot{\ifx\@let@token.\else.\null\fi\xspace}
\def\eg{\emph{e.g}\onedot} 
\def\ie{\emph{i.e}\onedot}

\makeatother

\title{Styleclone: Face Stylization with Diffusion Based Data Augmentation}
\author{
    Neeraj Matiyali\textsuperscript{\rm 1},
    Siddharth Shrivastava,
    Gaurav Sharma\textsuperscript{\rm 1}
}
\affiliations{
    \textsuperscript{\rm 1} Indian Institute of Technology, Kanpur \\
}

\usepackage{bibentry}

\begin{document}

\maketitle

\begin{abstract}  
We present \emph{StyleClone}, a method for training image-to-image translation networks to stylize faces in a specific style, even with limited style images.  Our approach leverages textual inversion and diffusion-based guided image generation to augment small style datasets. By systematically generating diverse style samples guided by both the original style images and real face images, we significantly enhance the diversity of the style dataset.  Using this augmented dataset, we train fast image-to-image translation networks that outperform diffusion-based methods in speed and quality. Experiments on multiple styles demonstrate that our method improves stylization quality, better preserves source image content, and significantly accelerates inference. Additionally, we provide a systematic evaluation of the augmentation techniques and their impact on stylization performance.  
\end{abstract}

\section{Introduction}
\label{sec:intro}

Image stylization has become a very popular area of research and application. Almost all image editing applications have some form of stylization built in. Stylizing faces is specially interesting for consumption in social media, as well as for use in creative applications, media and marketing. The task has seen rapid progress with two popular approaches, (i) that take a style image as reference \cite{gatys2016image,huang2017arbitrary}, and (ii) that are conditioned by text, \eg ``style this image as anime". These approaches are primarily based on conditional diffusion models \cite{ho2020denoising,rombach2022high} and their derivatives. Such stylization methods are generic, \ie they are made to do potentially any kind of stylization, based on the reference image or text prompt. They have also been adapted to learn new styles as concepts by giving a set of images of a style of interest that the user would like to learn to use techniques such as textual inversion \cite{gal2022image}, LoRA \cite{hu2021lora} and DreamBooth \cite{ruiz2023dreambooth}. In such a \emph{style cloning} scenario, the aim is to make a stylization method adapted to the particular style. 

Image synthesis with diffusion models is relatively slow due to multiple evaluations of a denoising neural network. While this is suitable for general-purpose tasks, it is inefficient for style-specific face stylization. We propose prioritizing faster inference by leveraging diffusion models for extensive training dataset augmentation and training a lightweight, style-specific image translation model.

\begin{figure}[t]
	\centering
	\includegraphics[width=0.9\linewidth]{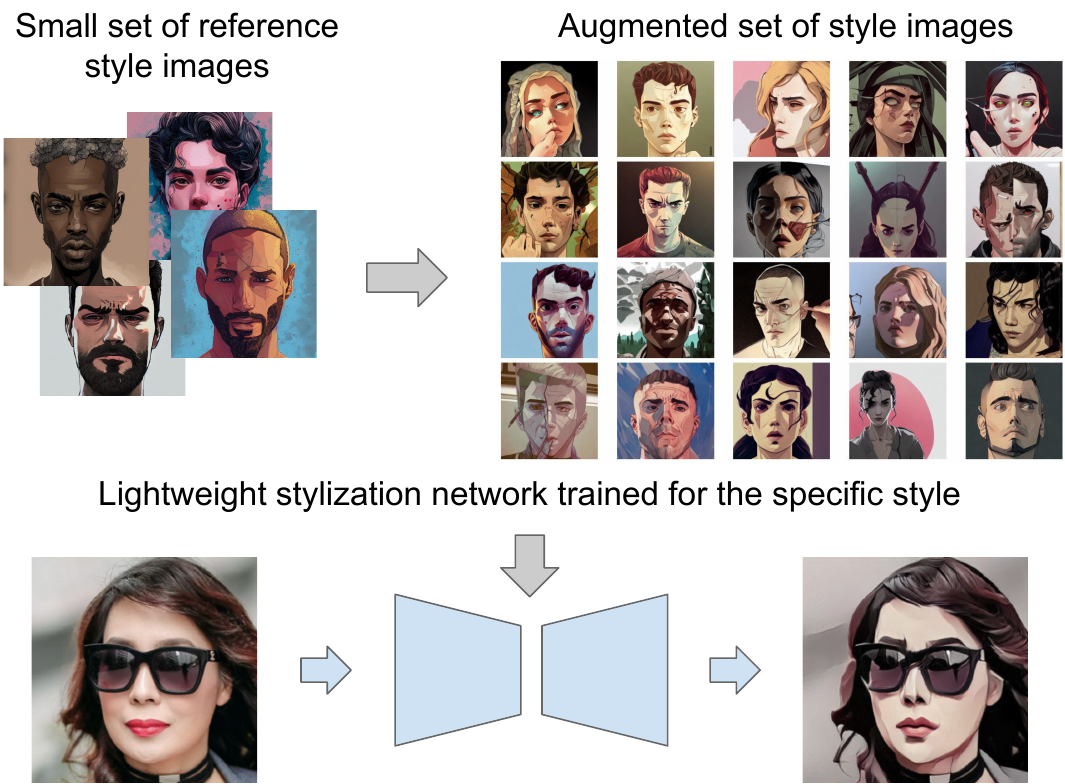}
	\caption{\emph{StyleClone} takes a small set of reference style images and augments them by generating similar styled images using textual inversion and diffusion models. It then trains a lightweight network with the augmented images which can mimic the style in the input reference images.}
	\label{fig:mainillus}
\end{figure}

We propose a method to train lightweight image translation network which is able to convert an input image into a particular style defined by a small set of images in that style. We propose to train such a network with adversarial training such that it learns to match the style of the output images to that of the style images. However, training it directly with the small set of reference style image does not give satisfactory results. Further, images generated by such a method carry strong distortions and deviations from the input image, and thus cannot be used directly as reasonable outputs. Moreover, diffusion based methods are heavy and slow, and thus not suitable for many applications that prioritize fast inference. Therefore, we propose an augmentation method, which takes the set of style images and learns to generate similar styled images using textual inversion \cite{gal2022image} and image-guided diffusion \cite{meng2021sdedit}. We show that when such a set is curated with a carefully designed augmentation strategy, it can be used to train lightweight image translation network with adversarial loss. We curate such a set by first using textual inversion to learn the concept represented by the style images. We then use it with diffusion model for guided image synthesis with the original input style images (self augmentation), as well as real face images (cross augmentation). While the augmentations generated with self augmentation closely match the input style images, and are more stable and less distorted, cross augmentation generates more varied and diverse images, albeit with some distortions.  Specifically, cross augmentation helps increase the coverage of style images in the augmented training data in terms of facial identity, head pose, colors, and background appearance.  It makes the trained translation network more robust to variations in the input face images and helps preserve the identity and structure of input faces better during stylization.  On the other hand, self augmentation, complements it by enforcing the constraint that the style of the final augmented training set is close to that of the input style images. Finally, we train the image translation network on the source domain consisting of real faces and the target domain consisting of augmented style images. The network is trained with an adversarial loss, without assuming any pairing between real and styled images. 

To summarize, our contributions are as follows. (i) We propose to use lightweight models for face stylization based on a small set of reference images in the style of interest. (ii) We use textual inversion and guided diffusion based data augmentation to generate examples that enable the training of the lightweight models. (iii) We propose two modes of augmentation (self and cross) based on the choice of guiding images and study their effect on the stylization results.  (iii) We compare our method with diffusion based image-to-image approaches, and perform various ablation experiments studying different aspects of the proposed method.

\section{Related Work}
\label{sec:rel-work}

\subsubsection{Image-to-Image Translation}
Image-to-image translation maps images from one domain to another, typically using adversarial training \cite{goodfellow2014generative}. Paired methods rely on aligned training data and combine adversarial and reconstruction losses \cite{isola2017image,wang2017high}, while unpaired methods like CycleGAN \cite{zhu2017unpaired} introduce cycle-consistency. CUT \cite{park2020contrastive} improves efficiency with contrastive loss. While fast at inference, these models require large training datasets. We mitigate this by first augmenting with diffusion models and then training a CUT-based stylization network.

\subsubsection{Portrait Stylization using StyleGAN}
StyleGAN \cite{karras2019style,Karras2019stylegan2} is widely used for portrait generation and stylization, often pretrained on FFHQ \cite{karras2019style}. Toonify \cite{pinkney2020resolution} adapts it for cartoon-like styles. StyleAlign \cite{wu2021stylealign} exploits latent space alignment, while Deceive D \cite{jiang2021deceive} and \cite{ojha2021few} propose data-efficient fine-tuning strategies. Image inversion is key for generalization \cite{richardson2021encoding,song2021agilegan}, enabling tasks like sketch-to-face or exemplar-based stylization \cite{yang2022pastiche}. However, inversion can limit robustness to out-of-distribution inputs. DCT-Net \cite{men2022dct} addresses this by avoiding inversion during inference.

\subsubsection{Diffusion-Based Stylization}
Diffusion models \cite{ho2020denoising,rombach2022high} have shown exceptional generative capabilities due to large-scale image-text training. Techniques like ControlNet \cite{zhang2023adding}, textual inversion \cite{gal2022image}, LoRA \cite{hu2021lora}, and DreamBooth \cite{ruiz2023dreambooth} enhance style control and adaptation. Inversion methods like DDIM \cite{preechakul2022diffusion,song2020denoising} enable editing and stylization \cite{cheng2023general,kim2022diffusionclip,zhang2023inversion}. Style transfer methods such as ZeCon \cite{yang2023zero} and attention-based approaches \cite{cao2023masactrl,chung2024style,deng2023z,liu2023portrait} improve controllability. However, slow sampling makes diffusion models less suited for speed-sensitive applications.

\section{Approach}
\label{sec:styleclone_approach}

\begin{figure*}[t]
	\centering
	\includegraphics[width=0.8\linewidth]{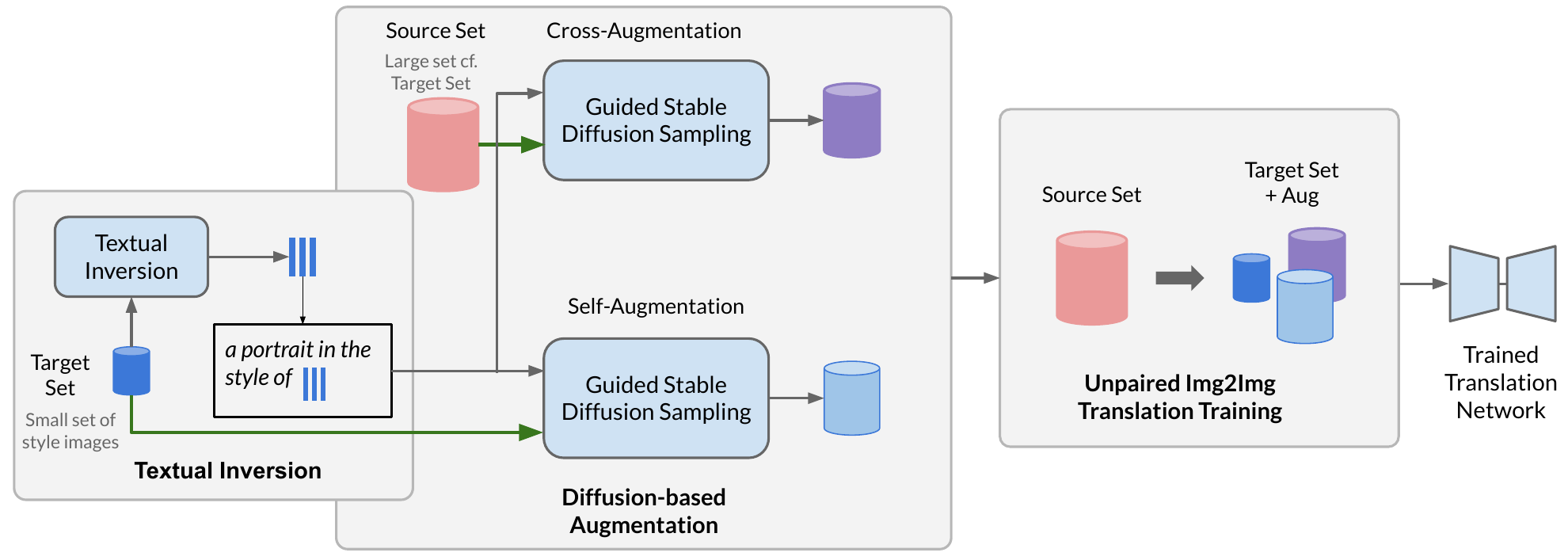}
    \caption{\textbf{Illustration of the proposed method.} The method works in following three steps. (i) It learns the style concept from the small set of target style images using textual inversion. (ii) It uses diffusion to augment the original style set (self augmentation), as well stylize a source set of real images (cross augmentation), to generate more stylized images. (iii) Finally it uses the augmented set to train a image stylization model using adversarial training.} 
	\label{fig:pipeline}
\end{figure*}


\subsection{Overview}

We address the task of face stylization with an image to image translation network. We consider two datasets $\mathcal{S}$ and $\mathcal{T}$, the source and the target respectively. $\mathcal{S}$ represents a large collection of real face images, and $\mathcal{T}$ represents a relatively much smaller set of images representing the style of interest. Both these datasets are independent, and do not contain any paired examples. A challenge while using traditional unpaired training strategies~\cite{park2020contrastive,zhu2017unpaired} is that the dataset $\mathcal{T}$, containing style images, is often not large enough to train an image translation model. To address this, we utilize pretrained text-to-image latent diffusion models \cite{rombach2022high} combined with textual inversion \cite{gal2022image}.

Our approach consists of three main steps (see Fig.~\ref{fig:pipeline}). (i) Utilizing textual inversion to acquire the style concept through learning style-specific textual embeddings. (ii) Employing the learned style embeddings to generate samples in the target style using a diffusion model. The style samples are generated with the guided image-to-image diffusion sampling \cite{meng2021sdedit} with the inputs from either the original style set $\mathcal{T}$ (self augmentation) or the source set $\mathcal{T}$ (cross augmentation). (iii) The final step is to train an image-conditioned generative model using unpaired samples from $\mathcal{S}$ and $\mathcal{T^+}$, where $\mathcal{T^+}$ is the target set augmented with the images generated in the second step.  

\subsection{Learning the Style with Textual Inversion}
The first step for enabling augmentation of the small set of style images is to learn to generate images in that style using a diffusion model. In order to generate samples in the style of $\mathcal{T}$ using text-to-image stable diffusion model, we need to describe the style in the input prompt. We achieve this with textual inversion \cite{gal2022image}. We use a placeholder token string $T_*$ to represent the style of the target set $\mathcal{T}$ that can be used inside the text prompts for the diffusion model. During the text embedding process of the diffusion models, the placeholder $T_*$ is replaced with a set of embedding vectors denoted by $t_*$. We sample text prompts from a set of simple template prompts such as ``A picture in the style of $T_*$" or ``A rendering in the style of $T_*$" and the style token embeddings $t_*$ are optimized to synthesize the images in $\mathcal{T}$.

After optimization, new samples can be generated in the style of $\mathcal{T}$ by using the learned token $T_*$ in the input prompts.


\subsection{Augmentation with Diffusion-based Guided Image Synthesis}

Image synthesis with a diffusion model is achieved by gradually denoising a random Gaussian noisy image. In the continuous formulation of diffusion models, the image $\mathbf x(t)$ (or its latent features in the case of latent diffusion models) goes from $t=1$ (pure Gaussian noise) to $t=0$ (with zero noise) by iteratively applying a denoising neural network according to a reverse stochastic differential equation (SDE). In text-to-image diffusion models, this denoising neural network is conditioned on an input text prompt that can control the content of the sampled image $\mathbf{x}(0)$.

As described in SDEdit \cite{meng2021sdedit}, a pretrained text-to-image diffusion model can also be guided with an image by initiating the denoising process with a noisy version of the guide image $\mathbf x_g(t_0)$ ($0 < t_0 < 1$) instead of pure random noise ($t_0=1$).
The influence of the guide image on the final generated image can be controlled by varying the initial noise level $t_0$, which we also refer to as the guidance factor, with high values of $t_0$ having low influence of the guide image on the final sampled image and vice-versa.

\begin{figure*}[t]
    \centering
    \includegraphics[width=0.8\textwidth]{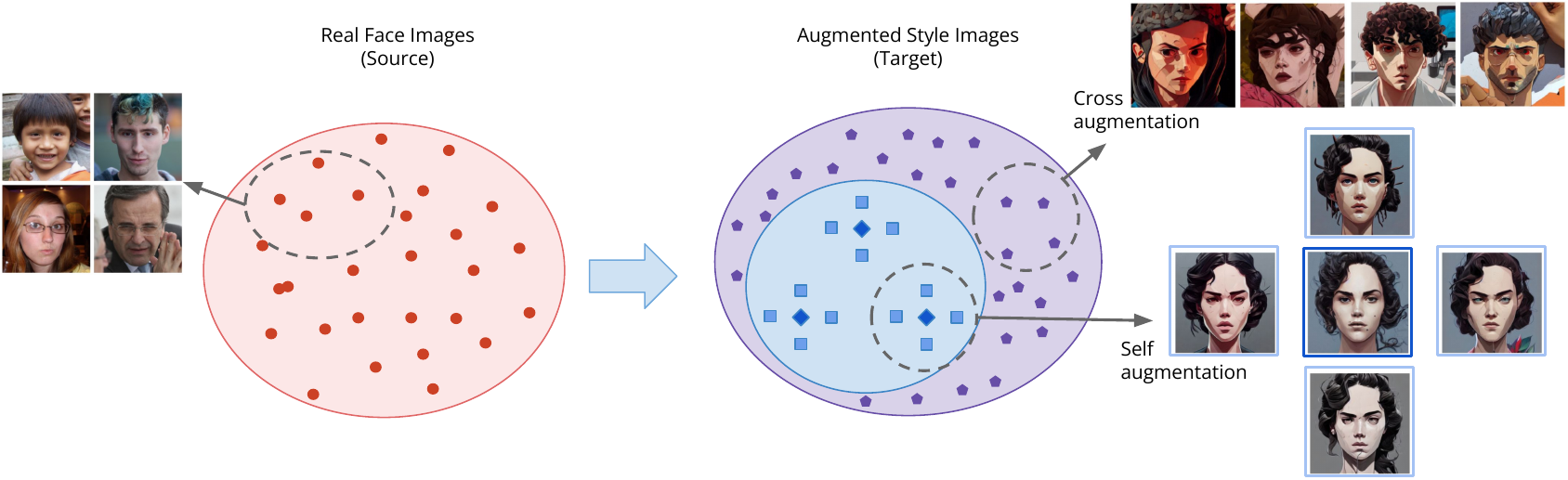}
    \caption{\textbf{Data augmentation.} We start with a small set of styled face images, and a large set of real face images and use diffusion with textual inversion to augment the styled image set. The styled images are augmented with diffusion model samples guided by the real images (cross augmentation) and the original style images (self augmentation).}
    \label{fig:data_aug}
\end{figure*}

The guided image synthesis provides two advantages. Firstly, by guiding on images with aligned and cropped faces (as in FFHQ dataset \cite{karras2019style}, for instance) we can encourage the generated images to also follow the same alignment and cropping conventions. Secondly, by taking the guiding images from the source set $\mathcal{S}$ or the style set $\mathcal{T}$, respectively referred to as cross and self augmentation (see Fig.~\ref{fig:data_aug}) we can generate two sets of augmenting datasets with distinct but complementary functions as described in the following subsections.

Let $\text{GIS}(x_g, t_0, c, r)$ denote the result of a guided image synthesis using \cite{meng2021sdedit} with $x_g$ as the guide image, $t_0$ as the guidance factor, and $r$ as the random seed. The image synthesis is conditioned on the input prompt $c$. We use a fixed prompt \textit{``a portrait in the style of $T_*$"} for all styles to avoid any manual intervention and keep the augmentation process automated. 

\subsubsection{Self Augmentation} 

\begin{figure}[t]
    \centering
    \includegraphics[width=0.8\columnwidth]{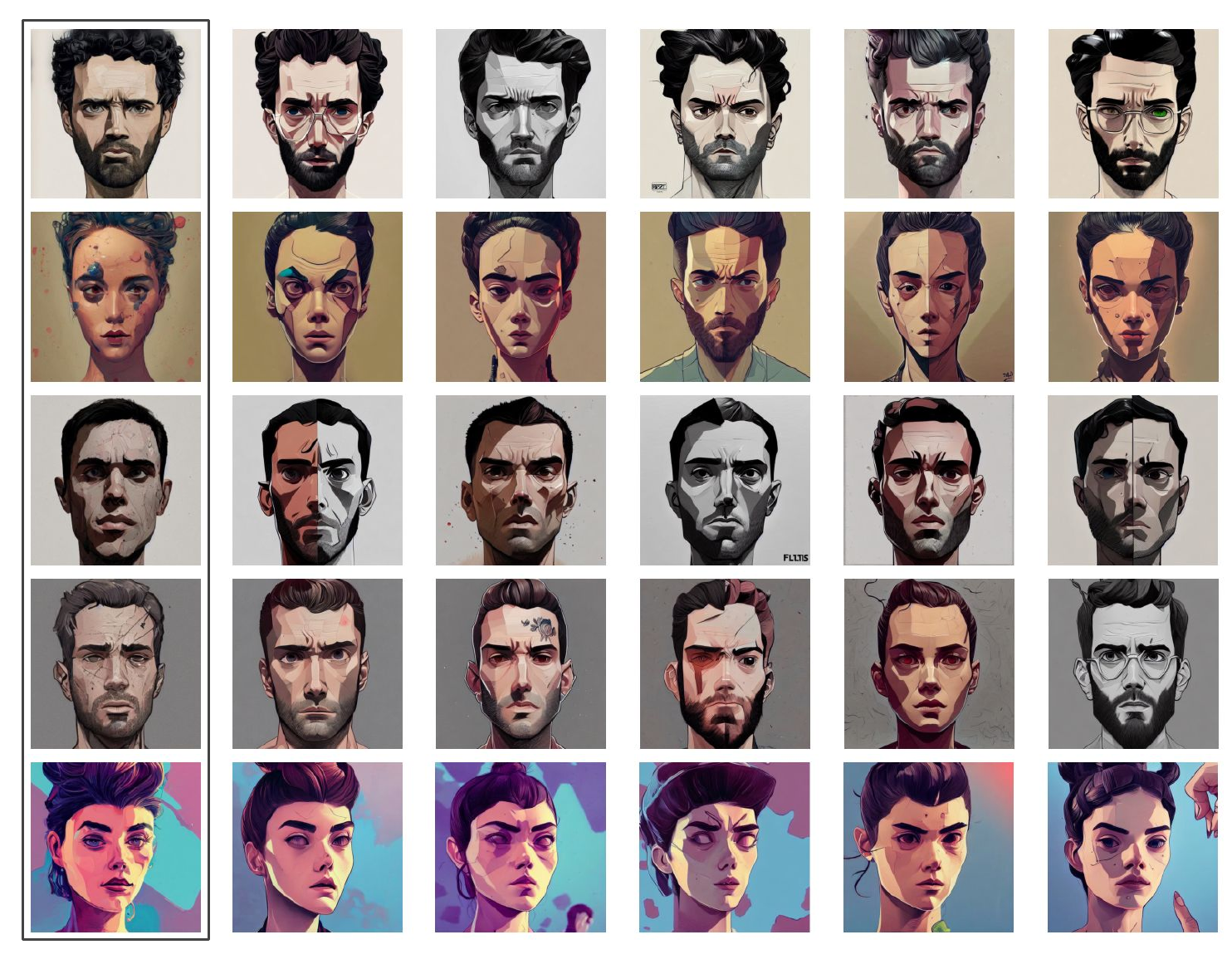}
    \caption{Self augmentation samples for the \textit{illustration} style.}
    \label{fig:self_aug_illus}
\end{figure}

In the self augmentation setup, we generate samples guided by the images from the original target set $\mathcal{T}$. We generate $N$ samples with the initial denoising time $t_0$,
\begin{equation}
	\mathcal{A}(\mathcal{T}, t_0) = \left\{ \text{GIS}(T_k, t_0, c, r_i) \mid k = i \bmod |\mathcal{T}| \right\}_{i=1}^N.
\end{equation}
Here, the guiding image $T_k \in \mathcal{T}$ is the $k$-th image in $\mathcal{T}$. We ensure a unique random seed for each generated image by using sample index as the random seed i.e. $r_i = i$.

A sample of generated images in this way with $t_0=0.8$ are shown in Fig.~\ref{fig:self_aug_illus} for the style \textit{illustration}.
The first column shows the guiding image and the remaining columns show samples with different random seeds. The overall style, layout, colors of the original images remains preserved while details of the face vary with different random seeds. Since the guiding image is from the same set which the token $T_*$ of the input prompt has been trained on, generated samples adhere closely to the original style of $\mathcal{T}$ while adding variety. However, this means that the new samples also inherit limitations of $\mathcal{T}$ especially the lack of diversity in aspects such as the head pose, background appearance, composition, color palette etc. Hence we also perform cross augmentation.


\subsubsection{Cross Augmentation}
\begin{figure}[t]
    \centering
    \includegraphics[width=0.8\columnwidth]{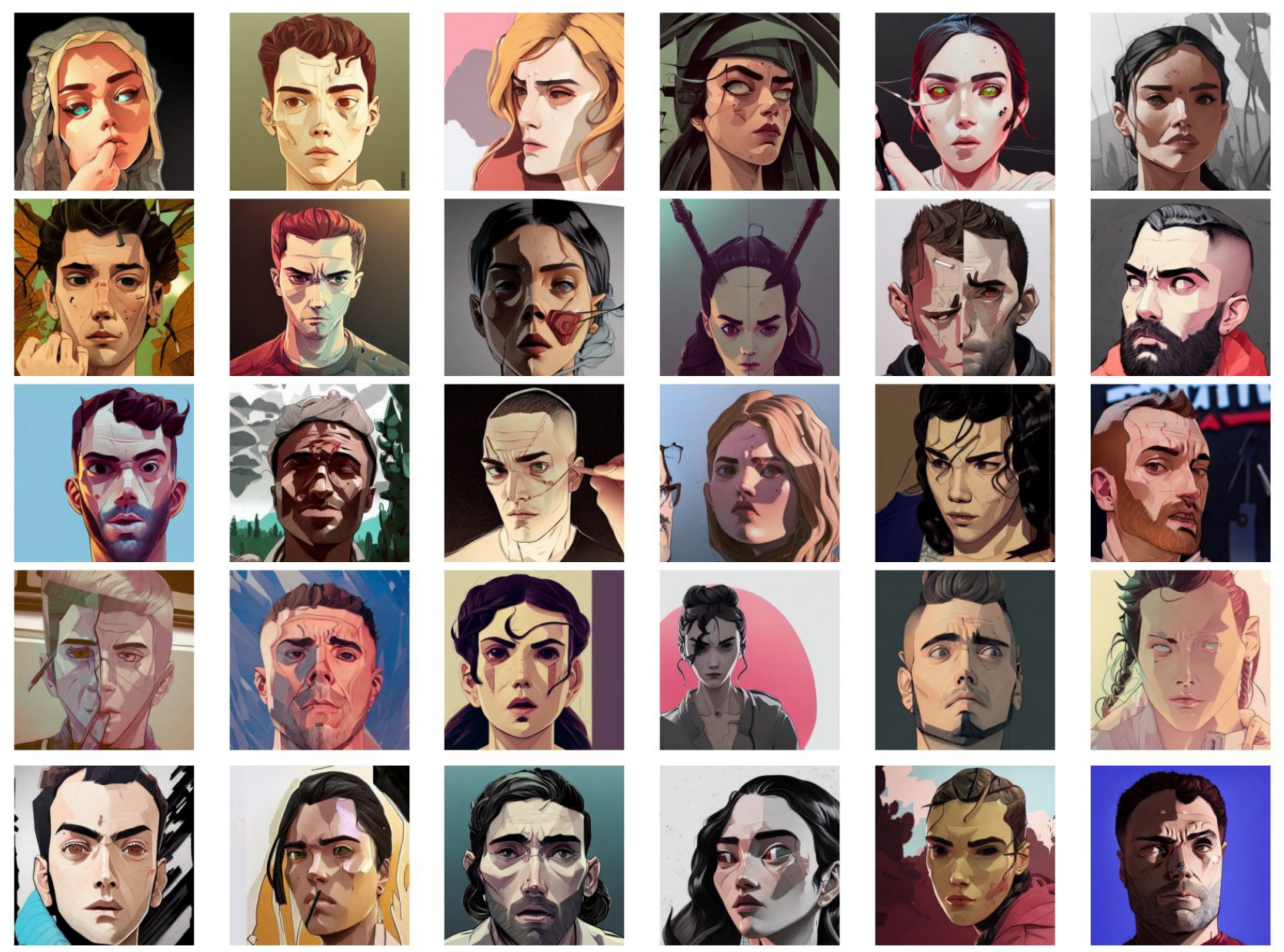}
    \caption{Cross augmentation samples for the \textit{illustration} style.} 
    \label{fig:cross_aug_illus_final}
\end{figure}

In this setup, we generate samples in a similar way as the self augmentation except that the guiding images are taken from the source set $\mathcal{S}$ of diverse real images. To generate $N$ samples with the guidance factor $t_0$
\begin{equation}
	\mathcal{A}(\mathcal{S}, t_0) = \left\{ \text{GIS}(S_k, t_0, c, r_i) \mid k = i \bmod |\mathcal{S}| \right\}_{i=1}^N
\end{equation}
Here, the guiding image $S_k \in \mathcal{S}$ is the $k$-th image in $\mathcal{S}$.

In Fig.~\ref{fig:cross_aug_illus_final}, we show a sample of generated images with different seeds for the \textit{illustration} style.
We observe that images generated with cross augmentation show a lot of diversity in terms of the pose, color palettes, appearances etc, as the guiding images are themselves quite diverse. However, they also exhibit artifacts and deviation from the original style due to the mismatch between the statistics of guiding images and the style set $\mathcal{T}$. 
As we increase $t_0$, the influence of guiding image decreases and the style resembles the target style more. Even with smaller values of $t_0=0.6$, the content eg.\ identity of the face, is significantly altered. We also tried experimenting
with $t_0 < 0.6$, but found that the outputs get too close to the guiding images and do not reflect the style in $\mathcal{T}$ to any meaningful degree.


\subsubsection{Final Augmented Target Set}
We combine samples generated with self augmentation and cross augmentation to get the final target set $\mathcal{T}^{+}$, which is union of $\mathcal{T}$, $\bigcup_{t_0\in T_\mathcal{T}} \mathcal{A}(\mathcal{T}, t_0)$, and $\bigcup_{t_0\in T_\mathcal{S}} \mathcal{A}(\mathcal{S}, t_0)$.
Here, $T_\mathcal{T}$ and $T_\mathcal{S}$ are the sets of guidance factors $t_0$ used for self and cross augmentation respectively.


\subsection{Training Image Translation Network}

As the final step of our pipeline, we train a lightweight image-to-image translation network with $\mathcal S$ and $\mathcal{T^+}$ as the source and the target training sets. The image translation network has an encoder-decoder UNet architecture which takes real face images as input and outputs an image in the desired style of $\mathcal{T}$.

Due to significant changes between the guiding images and the generated images in the cross augmentation process, especially for the higher values of $t_0$, we ignore any pairings between $\mathcal S$ and $\mathcal{T^+}$ and following CUT \cite{park2020contrastive} we train the stylization network in the unpaired setting. An adversarial loss is used to encourage the output to be indistinguishable from the samples in $\mathcal{T}$. A patchwise contrastive loss is used to maximize the mutual information between the input and output. We recommend readers to see \cite{park2020contrastive} for more details on training the image translation network.

\section{Experiments}
\label{sec:exps}


\subsection{Experimental Setup}
\label{sec:imp_details}

\noindent\textbf{Datasets.} We use FFHQ dataset of aligned and cropped face images as the source set $\mathcal{S}$.
It has 60000 train images and 10000 test images.
For the target style set $\mathcal{T}$, we use \textit{illustration}, \textit{impasto}, \textit{fantasy} and \textit{anime} style sets from DualStyleGAN~\cite{yang2022pastiche}, which contain 156, 120, 174 and 137 images respectively.
Ablation experiments are done on the $\textit{illustration}$ style.

\noindent\textbf{Implementation Details.} For the diffusion-based guided image synthesis \cite{meng2021sdedit,rombach2022high} and textual inversion \cite{gal2022image}, we use huggingface's diffusers library \cite{von-platen-etal-2022-diffusers}. We use Stable Diffusion v1.5\footnote{\url{https://huggingface.co/runwayml/stable-diffusion-v1-5}} as the fixed text-to-image latent diffusion model. In the textual inversion step, we learn 8 text embedding vectors for each style for 5000 iterations with a batchsize of 1. For self augmentation, we generate 10000 samples with a single guidance factor of $t_0 = 0.8$ i.e. $T_\mathcal T = \{0.8\}$. For cross augmentation, we generate 40000 samples conditioned on 10000 samples from the FFHQ dataset with the guidance factors in $T_\mathcal{S} = \{0.6, 0.7, 0.8, 0.9\}$. Thus, the final augmented set $\mathcal{T^+}$ has the $|\mathcal{T}| + 50000$ images. 
For image translation network, following \cite{park2020contrastive}, we use a Resnet based encoder-decoder architecture with 9 blocksk. The network is trained for 150000 iterations with a batch size of 4. We train the network on $256\times 256$ resolution.

\noindent\textbf{Evaluation.} We test stylization results from our method and baselines on a test set of 500 FFHQ test images. For quantitative comparison, we evaluate the FID~\cite{heusel2017gans,Seitzer2020FID} metric to assess the style of the translated results. 
The FID metric alone is not sufficient to evaluate the performance of an image translation network as it does not measure how well the content is preserved between the input and the stylized output. So, we also evaluate the average perceptual metric LPIPS~\cite{zhang2018perceptual} between the inputs and the outputs. By measuring both FID and LPIPS, we get a better insight into the trade-off between the content preservation and the stylization. 
When not mentioned explicitly, we measure FID against a self-augmented style set with samples generated with low guidance factor of $t_0 = 0.6$.
We use this set instead of the original style set $\mathcal T$, since $\mathcal T$ is not sufficiently large to be used as reference for FID computation.
Low guidance factor ensures that the reference set doesn't deviate much from the original style set.
For ablation experiments,  we also evaluate FID-S-0.8 and FID-C-0.8, which are FIDs evaluated against self and cross augmentation samples respectively, but with a higher guidance factor of $t_0=0.8$.


\subsection{Results}

\begin{figure}[t]
    \centering
    \includegraphics[width=\columnwidth]{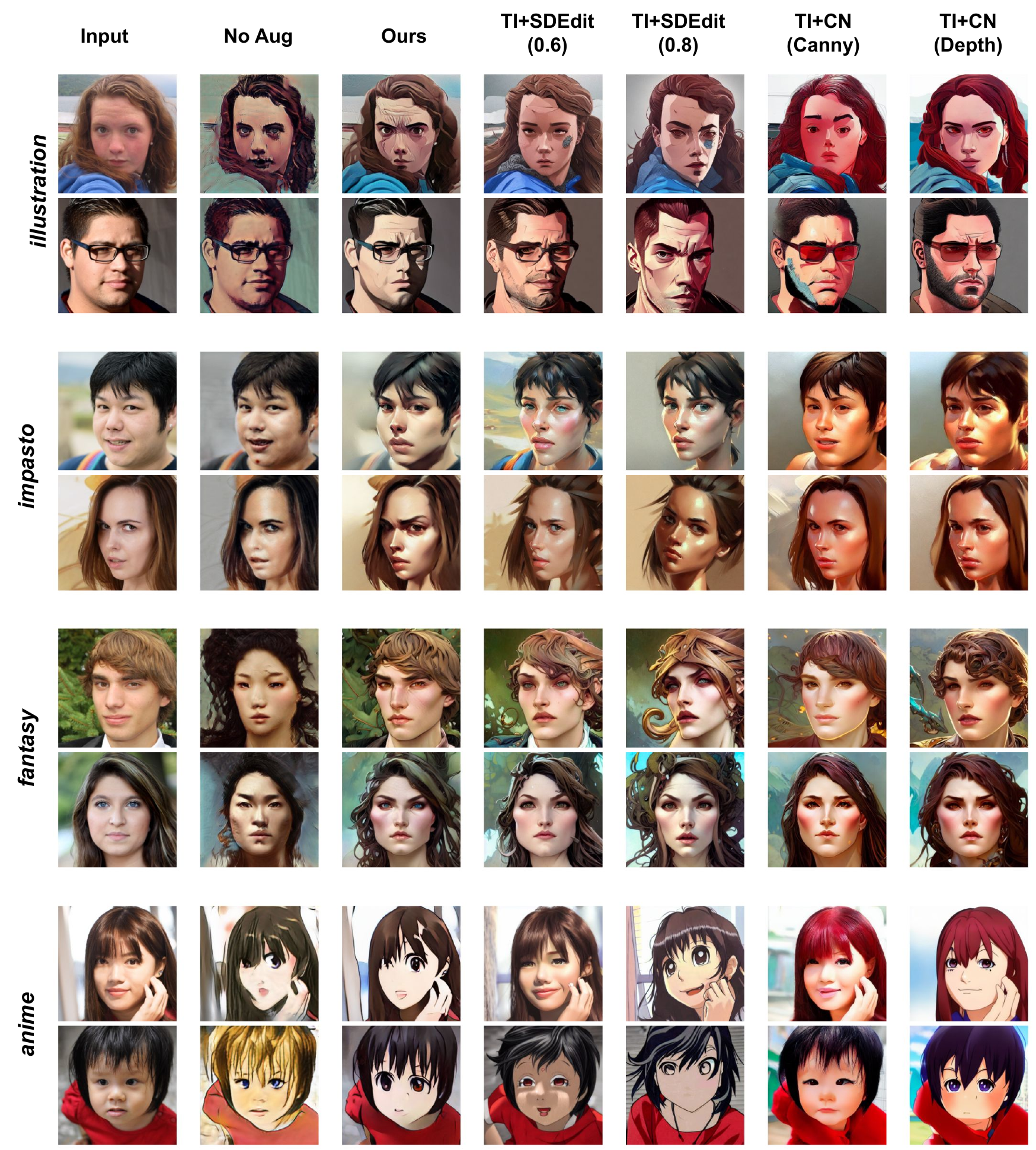}
    \caption{Comparison of image-to-image translation results with the baseline methods for different styles.}
    \label{fig:main_results}
\end{figure}

\begin{table*}[t]
    \centering
    \begin{tabular}{lrrrrrrrr}
        \toprule
		Style                                                                 & \multicolumn{2}{c}{illustration} & \multicolumn{2}{c}{impasto} & \multicolumn{2}{c}{fantasy} & \multicolumn{2}{c}{anime}                                                                                                             \\
		Method                                                                & \multicolumn{1}{c}{FID}          & \multicolumn{1}{c}{LPIPS}   & \multicolumn{1}{c}{FID}     & \multicolumn{1}{c}{LPIPS} & \multicolumn{1}{c}{FID} & \multicolumn{1}{c}{LPIPS} & \multicolumn{1}{c}{FID} & \multicolumn{1}{c}{LPIPS} \\ \midrule
		
		TI ~+~SDEdit  ($t_0=0.6$)  & \textbf{34.029}                           & 0.409                       & \textbf{35.056}                      & 0.358                     & \textbf{41.224}                  & 0.382                     & 144.901                 & 0.358                     \\
        TI ~+~SDEdit  ($t_0=0.8$)   & 34.223                           & 0.522                       & 56.810                      & 0.474                     & 65.115                  & 0.500                       & \textbf{56.167}                  & 0.559 \\ 
        TI ~+~CN  (Canny)      & 59.772  & 0.399 & 82.145  & 0.350  & 91.274  & 0.360  & 180.337 & 0.371 \\
TI ~+~CN  (Depth)      & 44.202  & 0.472 & 69.686  & 0.435 & 80.881  & 0.420  & 97.133  & 0.480 \\ \midrule 
No Aug                                                                & 137.943                          & 0.384                       & 154.445                     & \textbf{0.267}                     & 116.007                 & 0.429                     & 112.012                 & 0.534                     \\
		Ours                                                                  & 42.350                            & \textbf{0.345}                       & 61.704                      & 0.274                     & 84.713                  & \textbf{0.265}                     & 91.951                  & \textbf{0.325}                     \\ \bottomrule
    \end{tabular}
    \caption{Comparison of our stylization results with different methods on multiple styles. TI, SDEdit and CN refers to textual inversion \cite{gal2022image}, guided image generation \cite{meng2021sdedit} and ControlNet \cite{zhang2023adding} respectively.}
    \label{tab:my-table}

\end{table*}

\noindent\textbf{Comparison with baselines.} We compare our method with several baselines: (a) \textbf{NoAug}: image translation network trained with original target style set without any augmentation, (b) \textbf{TI+SDEdit} : SDEdit \cite{meng2021sdedit} with prompt containing text embeddings learned on the original style set via textual inversion \cite{gal2022image} 
(c) \textbf{TI+CN (Canny)}: Stable diffusion sampling conditioned on canny edges with ControlNet \cite{zhang2023adding} (d) \textbf{TI+CN (Depth)}: Stable diffusion sampling conditioned on depth maps with ControlNet. We use the same textual inversion based prompt for all diffusion based pipelines.

The qualitative comparison of our method with these baselines is shown in Fig.~\ref{fig:main_results}.
We observe that in the \textit{NoAug} variant, stylization networks fails to produce satisfactory results due to limited size of the target set for training. Besides not capturing the desired style \textit{NoAug} suffers from issues such as desaturated patches in outputs for the \textit{impasto} and significant change in the identity for the \textit{fantasy} style. These issues are eliminated by our augmentation strategy. Outputs from our method are able to capture the desired style well as well as preserve the identities and appearances of the input images. 

We find that while \textit{TI+SDEdit} baselines capture the target style really well, they do so at the expense of deviating too far from the input images. Even with the low guidance factor of 0.6 (higher influence of input image), the facial identities are changed significantly. In case of \textit{anime} style, we observe that \textit{TI+SDEdit (0.6)} fails to capture the style as the influence of real input is too high for exaggerated stylization necessary for the \textit{anime} style. 
With ControlNet (see \textit{TI+CN} columns in Fig.~\ref{fig:main_results}), some structural aspects of the input images eg. edges or depth are preserved depending on the conditioning signal, while other aspects such as color are not preserved. 
Our method in comparison preserves the input image's identity and appearance more effectively while having a significantly faster inference.

In Table~\ref{tab:my-table}, we present a quantitative comparison of our method against the baselines. The $\textit{NoAug}$ variant shows extremely high FIDs across all styles, which are significantly reduced by our augmentation framework, aligning with our observations from the qualitative results. Our method achieves the best LPIPS for three styles and remains close to $\textit{NoAug}$ on the \textit{impasto} style, demonstrating its strength in preserving input content. While SDEdit baselines, particularly with $t_0=0.8$, achieve better FIDs than our method, this comes at the cost of poor input content preservation, as evidenced by their high LPIPS. Except for the \textit{fantasy} style, our method outperforms both ControlNet baselines in FID while also achieving better LPIPS.


\subsection{Ablation Study}

\begin{figure}[t]
    \centering
    \includegraphics[width=0.7\linewidth]{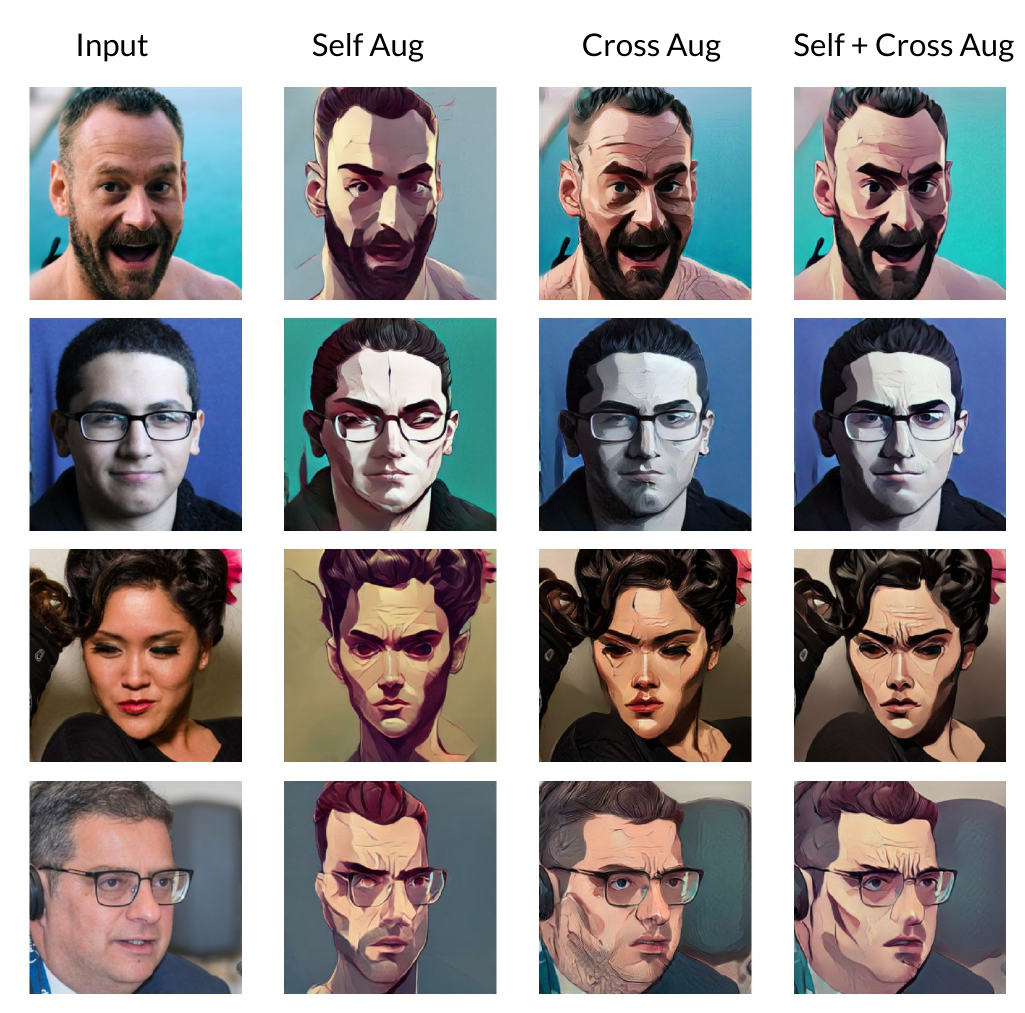}
    \caption{Comparison between self and cross augmentation for the \textit{illustration} style.}
    \label{fig:self_vs_cross_illus}
\end{figure}

\subsubsection{Impact of Self vs Cross Augmentation}

\begin{table}[t]
    \setlength{\tabcolsep}{0.7mm}
    \centering
    \begin{tabular}{lrrrr}
        \toprule
        \textbf{Model}   & \textbf{LPIPS}  & \textbf{FID-S-.6} & \textbf{FID-S-.8} & \textbf{FID-C-.8} \\ \midrule

        $K=0$ (NoAug)    & 0.51           & 163.3             & 177.0             & 163.2             \\
        $K=2500$         & 0.34           & 37.9              & 42.1              & 26.0              \\
        $K=10000$        & 0.33           & 36.8              & 41.7              & 25.0              \\
        $K=50000$ (Ours) & \textbf{0.32}  & \textbf{28.4}     & \textbf{33.0}     & \textbf{19.6}     \\ \midrule
        Self Aug Only    & 0.52           & \textbf{13.7}     & \textbf{12.1}     & 25.5              \\
        Cross Aug Only   & \textbf{ 0.31} & 34.22              & 40.4              & \textbf{25.2}     \\ \midrule
        CA@\{0.6, 0.7\}  & \textbf{0.31}           & 28.8              & 32.9              & 20.6              \\
        CA@\{0.8, 0.9\}  & 0.39  & \textbf{25.2}     & \textbf{27.6}     & \textbf{17.3}     \\

        \bottomrule
    \end{tabular}
    \caption{Quantitative comparison of translation results of different variants of our method for the \textit{illustration} style. LPIPS is measured between the FFHQ test images and the corresponding output of our translation networks. FID is measured against three sets of style samples generated with self-guidance and the cross-guidance, the same strategy used for training data augmentation.}
    \label{tab:quantitative_table}
\end{table}

To analyse the impact of self and cross augmentation, we train two additional variants: (a) in \textit{Self Aug}, the target set is only augmented with the images guided by style images of $\mathcal{T}$ while (b) in \textit{Cross Aug}, the target set is only augmented with the images guided by real images of FFHQ. The comparison is shown in Fig.~\ref{fig:self_vs_cross_illus}. We observe that without cross augmentation, the translation fails on some inputs due to lack of diversity in the training target set. For example, in the third row the identity of the face is drastically changed from input for \textit{Self Aug} or in the last row the deviation from the frontal head pose is not handled well.  This reflects the lack of images in the training set that are similar to those inputs. Similarly, in the last row, the output of \textit{Self Aug} shows an unnatural combination of the input pose and the frontal pose showing the bias towards the frontal pose in the self augmentation training samples. Both \textit{Cross Aug} and \textit{Ours} preserve the elements of input image better than the \textit{Self Aug} because of the diverse FFHQ-guided samples in the target training set. We find that, in general, we get a slightly greater degree of stylization with \textit{Ours} in comparison to the variant with cross augmentation only and some undesired artifacts are suppressed by also using self augmentation (see the neck region of \textit{Cross Aug} and \textit{Ours} in the fourth row).
In Table~\ref{tab:quantitative_table}, we show comparison of these variants in terms of quantitative metrics FID and LPIPS. First, we notice that \textit{Self Aug} has a significantly better FID than \textit{Self Aug} or \textit{Ours} (see \textit{FID-S-0.8} and  \textit{FID-S-0.6}: self augmented style images used as reference).  This is expected, considering that the reference used for FID computation uses the same strategy (guided generation with original style images) as the one used for generating the training images for \textit{Self Aug}. Second, \textit{Ours} performs better than \textit{Cross-Aug} for all FID metrics including \textit{FID-C-0.8} showing the importance of self-augmentation in helping the network produce results that are closer to the desired style. While the \textit{Self Aug} might be closer to the original style as measured by \textit{FID-S-0.8} and  \textit{FID-S-0.6}, they do not preserve the contents of the input images well as we can see from the LPIPS metric (also see Fig.~\ref{fig:self_vs_cross_illus}). LPIPS for \textit{Self-Aug} is much worse than \textit{Cross-Aug} and \textit{Ours}. Our final model that uses both self and cross augmentation provides a much better balance of content preservation (LPIPS) and stylization (FID) than either of the other two variants.

\subsubsection{Impact of Augmentation Size} 

\begin{figure}[t]
    \centering
    \includegraphics[width=0.8\linewidth]{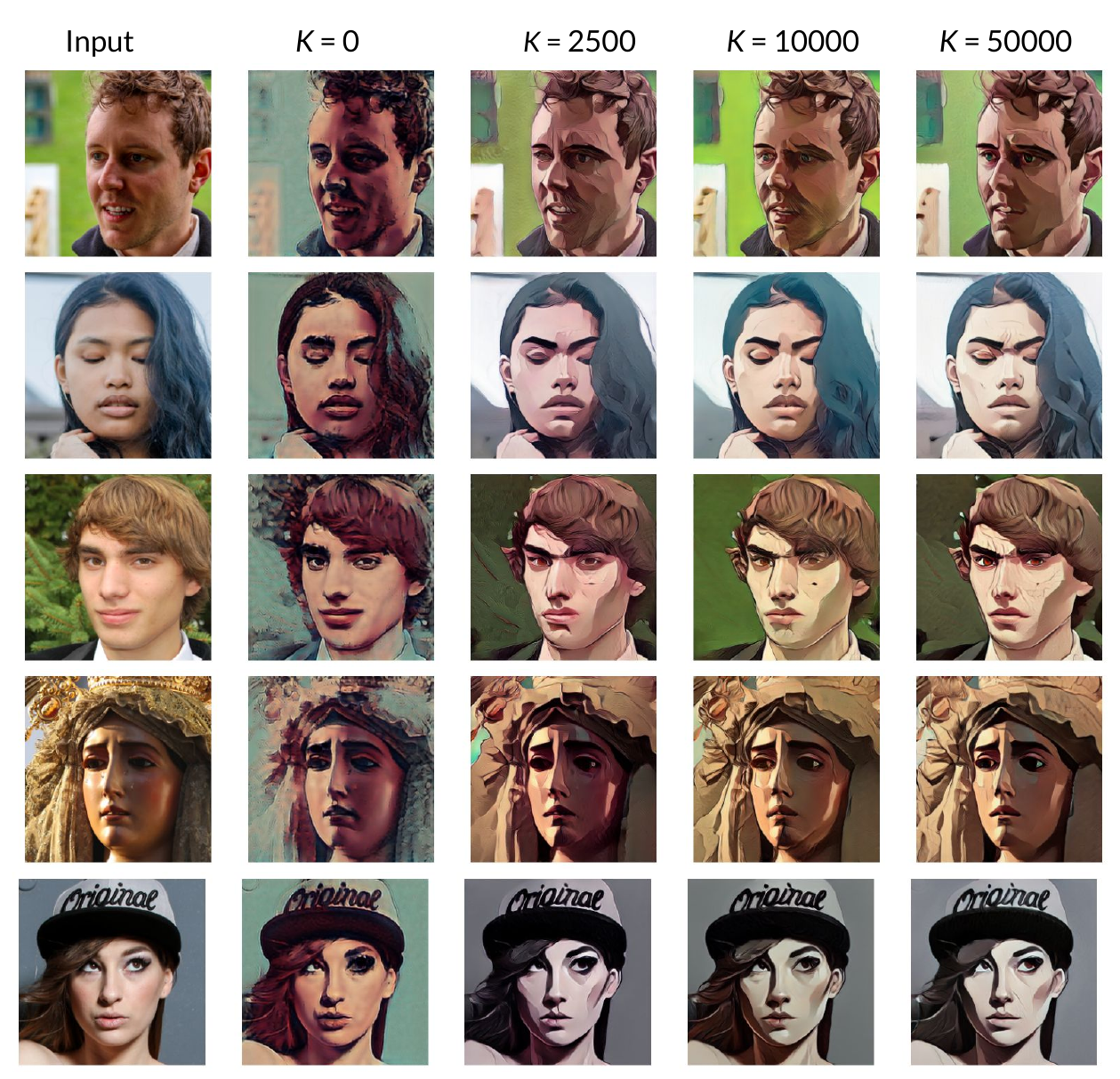}
    \caption{Effect of size of the augmented target set for the \textit{illustration} style.}
    \label{fig:aug_size_ablation}
\end{figure}

\begin{figure}[t]
    \centering
    \includegraphics[width=0.8\columnwidth]{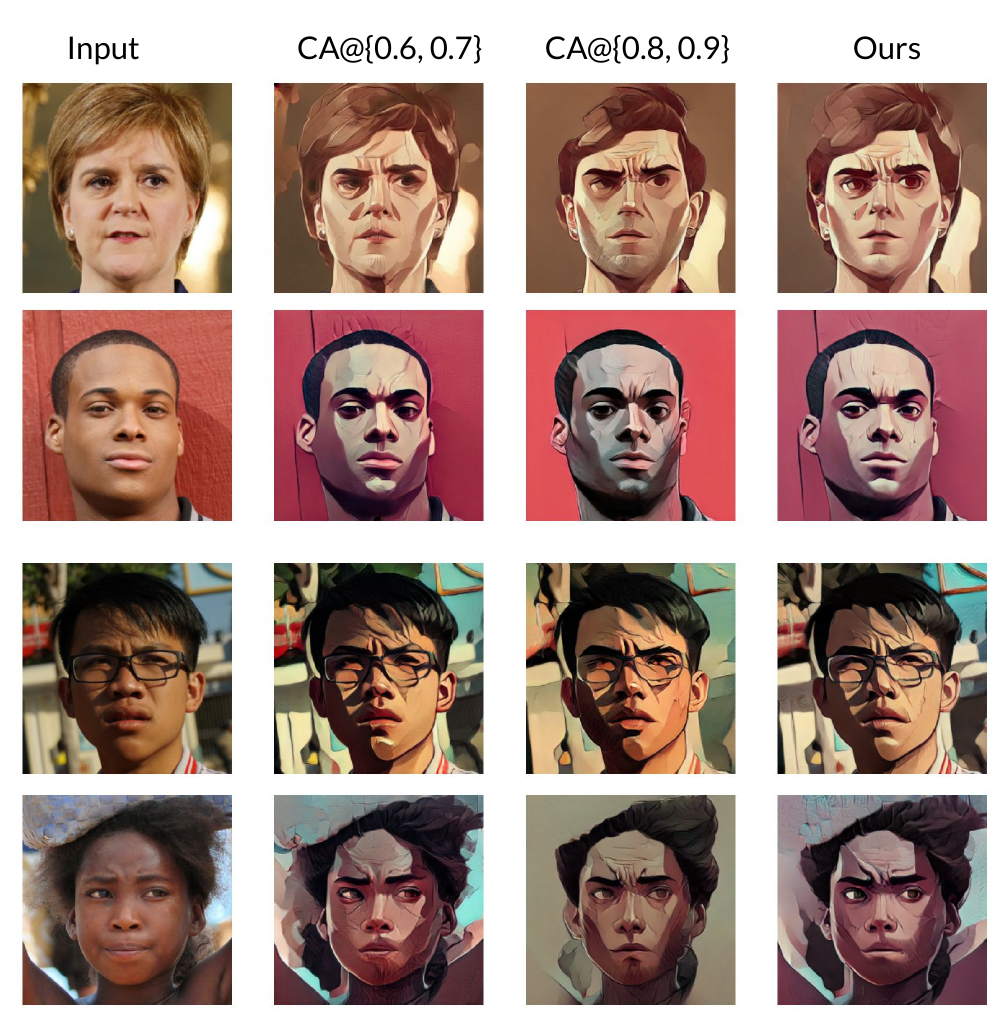}
    \caption{Effect of different guidance factors used for cross augmentation in the final translation results for the \textit{illustration} style.}
    \label{fig:noise_level_ablation}
\end{figure}

Let $\mathcal{A}$ be the set of all self and cross augmentation images combined. To investigate the effect of different augmentation sizes of the training target set, we take a random sample of $K$ images $\mathcal{A}_K \subset \mathcal A, K\in \{0, 2500, 10000, 50000\}$ and train the translation network on $(\mathcal S, \mathcal T \cup \mathcal A_K)$. Note that $K=0$ and $K=50000$ correspond to the variants $\textit{No Aug}$ and $\textit{Ours}$ respectively.
We show the comparison of different $K$ in Fig.~\ref{fig:aug_size_ablation}. We observe that even with $K=2500$ we get significantly better results than without any augmentation. For smaller values $K\in \{0, 2500\}$, we observe a consistent cast of reddish color in the output images, reflecting a bias of the relatively smaller target training sets. On the other hand, for larger values $K\in \{10000, 50000\}$, no such color bias is noticeable and the colors are more aligned with tthe source images.
In Table~\ref{tab:quantitative_table}, we show a quantitative comparison of models with different $K$ used during training. All metrics improve consistently as we increase the number of augmented samples in the training set. The results clearly show that the translation networks perform much better when trained with more images, improving both content preservation and stylization quality as measured by LPIPS and FID metrics respectively.


\subsubsection{Effect of guidance factors used for cross-augmentation}

We study the effect of guidance factors $t_0$ on the quality of cross augmentation. While our final model $\textit{Ours}$ uses $T_\mathcal S = \{0.6, 0.7, 0.8, 0.9\}$ for cross augmentation, we train two more variants by either only using lower guidance factors $\{0.6, 0.7\}$ or higher guidance factors $\{0.8, 0.9\}$. We show the comparison in Fig.~\ref{fig:noise_level_ablation}. We observe that with smaller guidance factors (i.e. more influence of FFHQ real images on the target training set) the elements of input image are preserved better through the translation. For example, in the first row, the identity is preserved more with lower values of $t_0$ in comparison with the other ones. Similarly, in all rows the background details are preserved more with $t_0\in\{0.6, 0.7\}$ than $t_0\in\{0.8, 0.9\}$. This comparison demonstrates the importance of low values of $t_0$ during augmentation and also highlights the role of cross augmentation in general. The LPIPS and FID results in Table~\ref{tab:quantitative_table} confirm these observations. 
More influence of source real images during training data augmentation helps in improving the LPIPS distance between the input and output without sacrificing the quality of stylization as measured by FID. For $t_0\in\{0.6, 0.7\}$, we get a similar trade-off between LPIPS and FID as our final model with better LPIPS but slightly worse or similar FID. With $t_0\in\{0.8, 0.9\}$, we get better FIDs but with a significant increase in LPIPS which could be unacceptable for certain real-world applications.

\section{Conclusions and Future Works}

We presented a lightweight and fast network for face image stylization that requires only a small set of target style images. Our approach trains an image-to-image translation network by augmenting the style set using diffusion-based methods, incorporating textual inversion and conditional diffusion for image generation. Diffusion methods, however, face two key issues: (i) significant deviations from the input image with noticeable distortions, and (ii) heavy computational requirements leading to slow performance.

Our proposed method is tailored for a specific style, achieving orders-of-magnitude faster performance while producing higher-quality stylized images with improved content preservation compared to diffusion methods. This speed enables real-time and video applications for face stylization. As future work, we aim to extend this approach to video stylization, addressing challenges such as maintaining temporal consistency.

\bibliographystyle{aaai2026}
\bibliography{refs}

\end{document}